\documentclass[10pt,twocolumn,letterpaper]{article}

\usepackage{cvpr} \usepackage{times} \usepackage{epsfig} \usepackage{graphicx}
\usepackage{amsmath} \usepackage{amssymb}

\usepackage{makecell}

\usepackage[group-separator={,}]{siunitx}
\usepackage{booktabs}\usepackage{booktabs}
\usepackage{tabularx}
\usepackage[caption=false,font=footnotesize,labelfont=sf,textfont=sf]{subfig}

\usepackage[table]{xcolor}

\usepackage{hyperref}

\cvprfinalcopy

\graphicspath{{figures/}}

\ifcvprfinal\fi
\begin{document}

\title{Supersaliency: A Novel Pipeline for Predicting Smooth Pursuit-Based 
	Attention Improves Generalizability of Video Saliency}

\author{Mikhail Startsev, Michael Dorr\\ Technical University of Munich\\
{\tt\small \{mikhail.startsev, michael.dorr\}@tum.de}
}

\maketitle

\begin{abstract} 
	Predicting attention is a popular topic at the
	intersection of human and computer vision.
	However, even though most of the available
	video saliency data sets and models claim to target human observers'
	fixations, they fail to
	differentiate them from smooth pursuit (SP), a major 
	eye movement type
	that is unique to perception of dynamic scenes.
	In this work, we highlight the importance of SP and its prediction (which we call supersaliency, 
	due to greater selectivity compared to fixations), and aim to make its distinction 
	from fixations explicit for computational models. To this end,
	we (i) use algorithmic and manual annotations 
	of SP and fixations for two well-established video saliency data sets,
	(ii) train Slicing Convolutional Neural Networks for
	saliency prediction on either
	fixation- or SP-salient locations, and (iii) evaluate our and 26
	publicly available dynamic saliency models on three data sets against traditional saliency and supersaliency ground truth.
	Overall, our models outperform the state of the art in both the new supersaliency and the traditional saliency problem settings, for which literature models are optimized.
	Importantly, on two independent data sets, our supersaliency model shows greater generalization ability and outperforms all other models, even for fixation prediction.
\end{abstract}

\section{Introduction}

prediction has a wide variety of applications, be it in computer vision,
robotics, or art \cite{borji2013state}, ranging from
image and video compression \cite{guo2010novel, hadizadeh2014saliency} to such high-level tasks
as video summarisation \cite{marat2007video}, scene recognition \cite{siagian2007rapid}, or
human-robot interaction \cite{nagai2009computational}.
Its underlying idea is that in order to efficiently use the limited neural bandwidth,
humans sequentially sample informative parts of the visual input with the high-resolution
centre of the retina, the \emph{fovea}. The prediction of gaze should thus be related to
the classification of informative and uninformative video regions.
However, humans use two different processes to foveate visual content. During fixations,
the eyes remain mostly stationary; during smooth pursuit (SP), in contrast, a moving target 
is tracked by the eyes to maintain foveation. Notably, SP is impossible without
such a target, and it needs to be actively initialized and maintained.
For models of attention, this is a critical distinction:
Because the eyes are stationary (``fixating'') in their default state,
``spurious'' fixations may be detected even if a subject is not attentively
looking at the input; SP, however, always co-occurs with attention.
In addition,
visual sensitivity seems to be improved during SP (e.g.\ higher chromatic contrast
sensitivity \cite{schutz2011eye} and enhanced visual motion prediction \cite{spering2011keep}).

\begin{figure}[!t]
	\centering
	\includegraphics[width=0.6\linewidth]{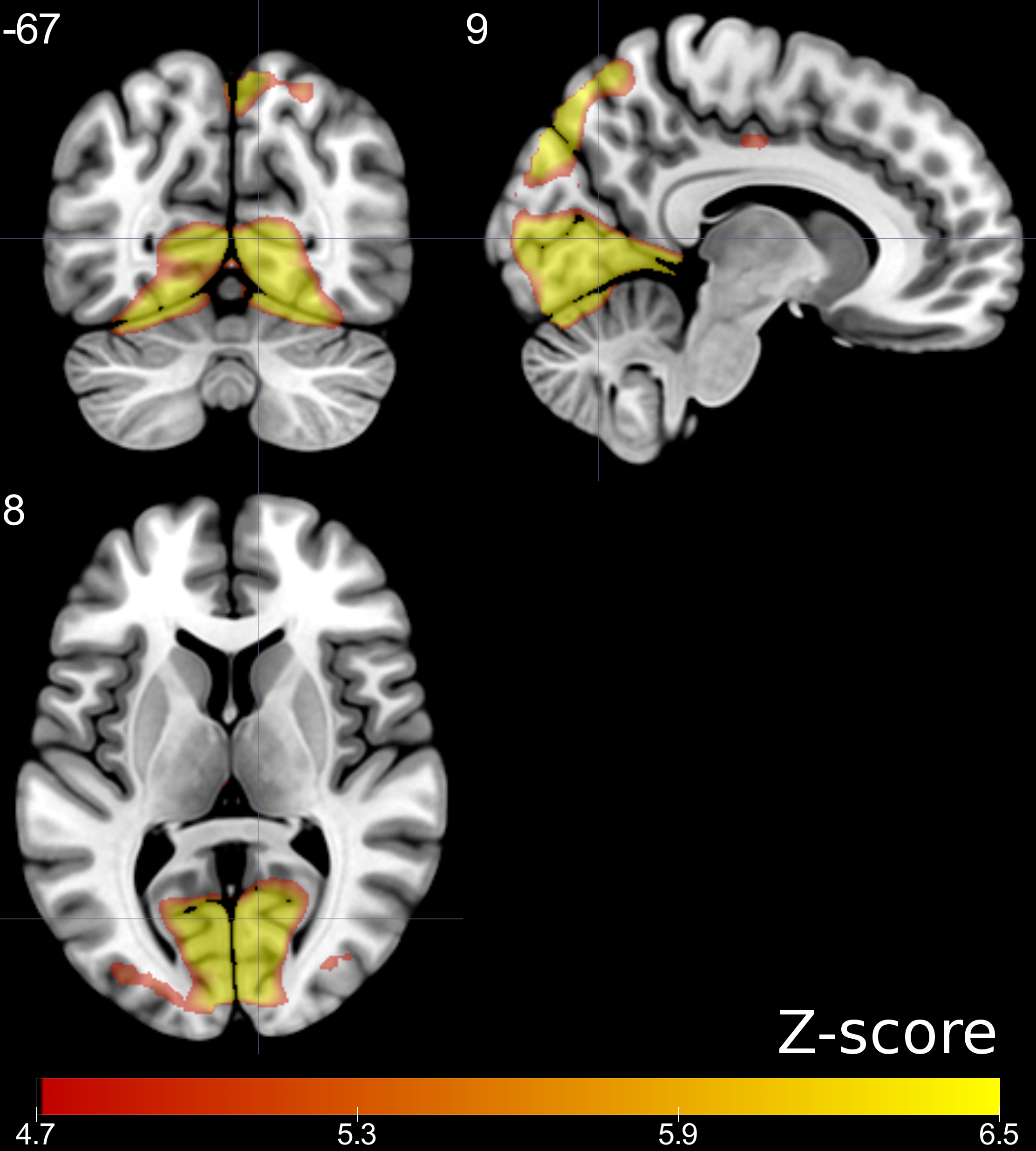}%
	\caption{Empirically observed neurological differences between fixation and smooth pursuit: Large brain areas (highlighted) show significantly increased activation levels during pursuits compared to fixations (detected by \cite{agtzidis2016smooth}) in the \textit{studyforrest} data set \cite{hanke2016studyforrest}; none demonstrate the inverse effects.}
	\label{fig:sp_vs_fix:brain}
\end{figure}

In practice, it is difficult to segment the -- often noisy -- eye tracking signals
into fixations and SP, and thus many
researchers combine all intervals where the eyes are keeping track of a point or an object into ``fixations'' \cite{pelz2001oculomotor}.
Nevertheless, it is well-established that e.g.\ individuals with schizophrenia show altered
SP behaviour \cite{sereno1995antisaccades,silberg2017free}, and recently new methods
for gaze-controlled user interfaces based on SP have been presented \cite{vidal2013pursuits, esteves2015orbits, schenk2016spock}.

The ultimate goal of all eye movements and perception, however, is to facilitate action in the real world. 
In a seminal paper \cite{land2006eye}, Land showed that gaze strategies, and SP in particular,
play a critical role during many everyday activities. Similar results have been found 
for driving scenarios, where attention is crucial.
Studies show that tangential \cite{authie2011optokinetic}
and target \cite{lappi2013eye} locations during curve driving are ``fixated''
with what actually consists, in part, of SP. In natural driving, roadside
objects are often followed with pure SP, without head motion
\cite{lappi2017systematic}.
Following objects that are moving relative to the car with gaze
(by turning the head, via an SP eye movement, or a combination of both) is a
clearer sign of attentive viewing, compared to the objects of interest crossing
the line of sight. 

\begin{figure}[!t]
	\centering
	\includegraphics[width=\linewidth]{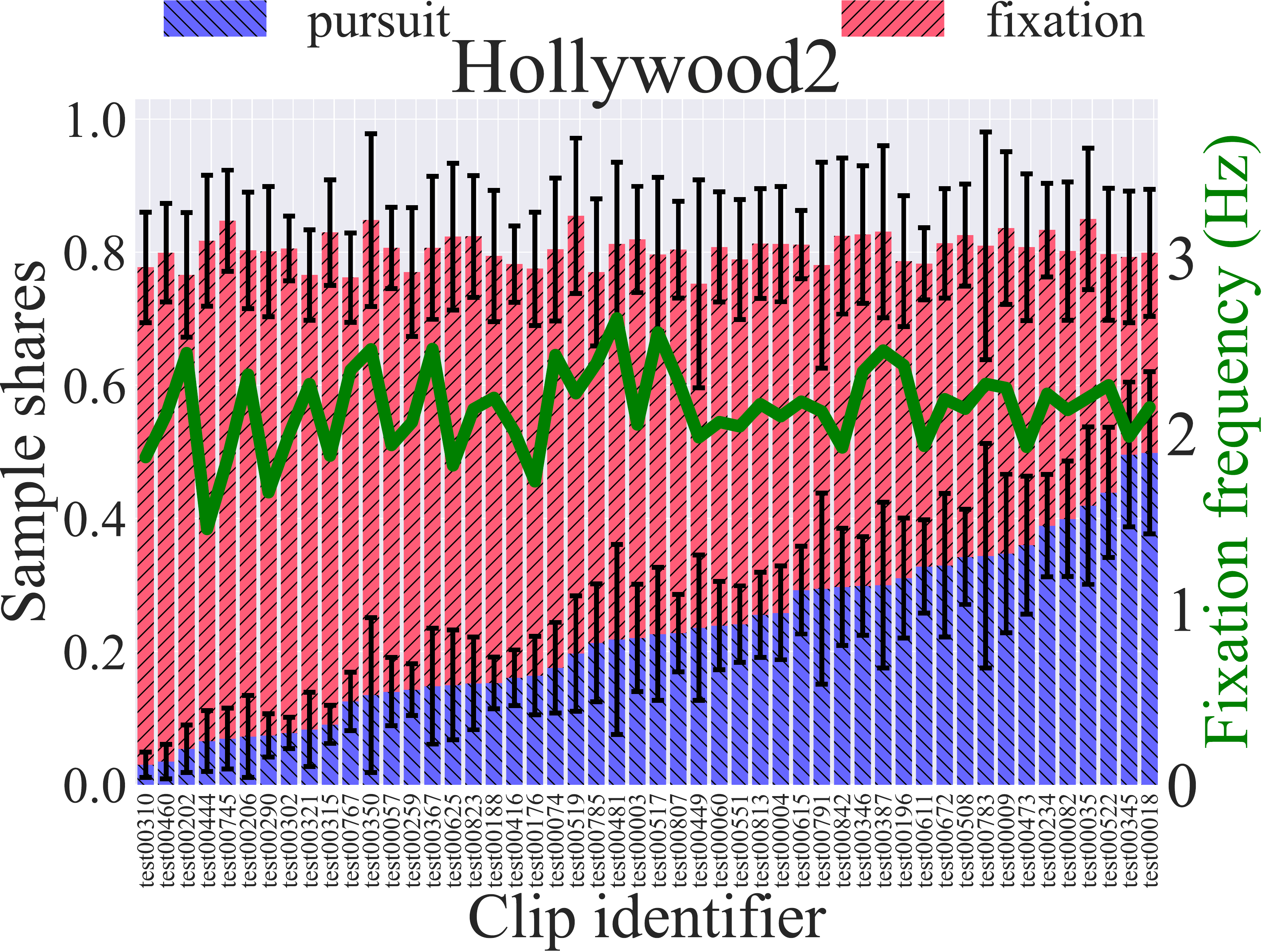}%
	\caption{Behavioural differences between fixation and smooth pursuit: Saliency metrics typically evaluate against fixation onsets, which, as detected by a traditional approach \cite{dorr2010variability} (green line), are roughly equally frequent across videos. 
		However, applying a more principled approach to separating smooth pursuit from fixations \cite{agtzidis2016smooth} reveals great variation in the number of fixation (red bars) and pursuit (blue bars) samples (remaining samples are saccades, as well as blinks and other unreliably tracked samples).}
	\label{fig:sp_vs_fix:shares}
\end{figure}

\figurename~\ref{fig:sp_vs_fix:brain} and \figurename~\ref{fig:sp_vs_fix:shares} show two analyses corroborating the importance of
SP for models of attention for the more tractable 
task of video watching. In \figurename~\ref{fig:sp_vs_fix:brain}, data from the
publicly available \textit{studyforrest} data set\footnote{This data was 
	obtained from the OpenfMRI database. Its accession number is ds000113d.}
\cite{hanke2016studyforrest}, which combines functional brain imaging and eye
tracking during prolonged movie watching, were comparatively evaluated for SP
vs.\ fixation episodes. The highlighted voxels show that large brain areas are
more active during SP compared to fixations; notably, no brain areas were more
active during fixation than during SP. In other words, SP is representative of
greater neurological engagement. 
The sparser selectivity of SP is demonstrated in
\figurename~\ref{fig:sp_vs_fix:shares}, where the relative share of SP and fixation
gaze samples is plotted for 50 randomly selected clips from Hollywood2.
Even though the number of traditionally detected fixations (but not their duration) is 
roughly the same for all clips, the amount of SP ranges from almost zero to
half of the viewing time.

Taken together, these observations let us hypothesize that SP is used
to selectively foveate video regions that demand greater cognitive resources,
i.e.\ contain more information.
In practice, automatic pursuit classification as applied to the 
\textit{studyforrest} and Hollywood2 data sets may not be perfect,
but the results in \figurename~{\ref{fig:sp_vs_fix:brain}} corroborate 
that even with potentially noisy detections, 
SP corresponds to higher brain activity, and thus to more meaningful saliency.

Explicitly modelling SP in a saliency pipeline should benefit the classification
of informative video regions. Beyond a better understanding of attention, there
might also be direct applications of SP prediction itself, e.g.\
in semi-autonomous driving (verification of attentive supervision),
telemedicine (monitoring of SP impairment as a
vulnerability marker for schizotypal personality disorder
\cite{siever1984impaired}, e.g.\ during TV or movie watching \cite{silberg2017free}), or gaze-based interaction (analysis
of potential distractors in user interfaces for AR/VR).

Despite the fundamental differences between SP and fixations, however, available
data sets ignore this distinction,
and saliency models naturally follow suit
\cite{leboran2017dynamic, khatoonabadi2015bits}.
In fact, not one of the video saliency models we came across mentions the tracking of objects 
performed via SP, and the only data set we found 
to purposefully attempt separating SP from fixations is GazeCom \cite{dorr2010variability}, which
simply discarded (likely) pursuits in order to achieve cleaner fixations.

In this manuscript, we extend our previous work \cite{startsev2018increasing} and make the following contributions: First, we introduce the problem of smooth pursuit prediction -- \textit{supersaliency}, so named due to the properties separating it from traditional, fixation-based saliency (e.g.\ see \figurename~{\ref{fig:sp_vs_fix:brain}} and \figurename~{\ref{fig:sp_vs_fix:shares}}). In this problem setting, the saliency maps correspond to how likely an input video location is to induce SP.
We then provide automatically labelled \cite{agtzidis2016smooth}, large-scale training and test sets for this problem (building on the Hollywood2 data set \cite{mathe2012dynamic}), as well as a manually labelled, smaller-scale test set of more complex scenes in order to test the generalizability of saliency models (building on the GazeCom data set \cite{dorr2010variability, sp-detection-site}). 
For both, we provide SP-only and fixation-only ground truth saliency maps.
We also discuss the necessary adjustments to the evaluation 
of supersaliency due to its high inter-video variance, introducing weighted averaging of individual clip scores.

Furthermore, we propose a deep dynamic saliency model for (super)saliency prediction, which is based on the slicing convolutional neural network (S-CNN) architecture \cite{shao2016slicing}. 
After training our proposed model for both saliency and supersaliency prediction on the same overall data set, we demonstrate that our models excel at their respective problems in the test subset of the large-scale data set, compared to two dozen literature models. Finally, we show that training for predicting smooth pursuit reduces data set bias: The supersaliency-trained model generalizes better to two independent sets (without any additional training) and performs best even for (traditional) saliency prediction. 

\section{Related work}

Predicting saliency for images has been a very active research field.
A widely accepted benchmark is represented by the MIT300 data set \cite{judd2012benchmark, mit-saliency-benchmark},
which is currently dominated by deep learning solutions.
Saliency for videos, however, lacks an established benchmark. 
It generally is a challenging problem, 
because, in addition to larger computational cost, 
objects of interest in a dynamic scene 
may be displayed only for a limited time and in different positions and contexts, 
so attention prioritisation is more crucial.

\subsection{Saliency prediction}

A variety of algorithms has been introduced to deal with human attention prediction \cite{borji2013state}.
Video saliency approaches broadly fall into two groups: 
Published algorithms mostly operate either in the original pixel domain
\cite{harel2007graph, guo2010novel, leboran2017dynamic, wang2017fixation}
and its derivatives (such as optic flow \cite{wu2016video} 
or other motion representations \cite{zhai2006visual}), 
or in the compression domain \cite{khatoonabadi2015bits, li2016fast, xu2017learning}.
Transferring expert knowledge from images to videos in terms of saliency prediction
is consistent with pixel-domain approaches, and the mounting evidence that motion
attracts our eyes contributed to the development of compression-domain algorithms.

Traditionally, from the standpoint of perception, 
saliency models are also separated into two categories based on 
the nature of the features and information they employ.
Bottom-up models focus their attention (and assume human observers do the same) 
on low-level features such as luminance, contrast, or edges. For videos, 
local motion can also be added to the list, and with it the video encoding information.
Hence, all the currently available compression-domain saliency predictors are 
essentially bottom-up.

Top-down models, on the contrary, use high-level, semantic information, such as 
concepts of objects, faces, etc. These are notoriously hard to formalize.
One way to do so would be to detect certain objects in the video scenes,
as was done in \cite{mathe2012dynamic}, where whole human figures, faces, 
and cars were detected. Another way would be to rely on developments in
deep learning and the field's endeavour to implicitly learn important
semantic concepts from data. In \cite{chaabouni2016deep}, either RGB space
or contrast features are augmented with residual motion information to account for the dynamic
aspect of the scenes (i.e.\ motion is processed before the CNN stage in a handcrafted fashion). 
The work in \cite{bazzani2017recurrent} uses a 3D CNN
to extract features, plus an LSTM network to expand the temporal span
of the analysis. Other researchers use further additional modules, 
such as the attention mechanism \cite{wang2018revisiting} or object-to-motion 
sub-network \cite{jiang2018deepvs}.

While using a convolutional neural network in itself
does not guarantee the top-down nature of the resulting model, its multilayer
structure fits the idea of hierarchical computation
of low-, mid-, and high-level features. A work by Krizhevsky et al.\ \cite{krizhevsky2012imagenet}
pointed out that whereas the first convolutional layer learned fairly simplistic 
kernels that target frequency, orientation and colour of the input signal,
the activations in the last layer of the network corresponded to a feature space,
in which conceptually similar images are close, regardless of the distance in 
the low-level representation space. 
Another study \cite{cadieu2014deep} concluded that, just like certain neural populations
of a primate brain, deep networks trained for object classification create such
internal representation spaces, where images of objects in the same category
get similar responses, whereas images of differing categories get dissimilar ones.
Other properties of the networks discussed in that work indicate potential
insights into the visual processing system that can be gained from them.

\subsection{Video saliency data sets}

A broad overview of existing data sets is given in \cite{winkler2013overview}.
Here, we dive into the aspect 
particularly relevant to this study -- the identification of ``salient'' locations of the videos, 
i.e.\ how did the authors deal with dynamic eye movements.
For the most part, this issue is addressed inconsistently. The majority of the data sets
either make no explicit mention of separating smooth pursuit from fixations (ASCMN \cite{riche2013dynamic}, SFU \cite{hadizadeh2012eye},
two Hollywood2-based sets \cite{mathe2012dynamic, vig2012space}, DHF1K \cite{wang2018revisiting}) or rely on the event detection 
built into the eye tracker,
which in turn does not differentiate SP from fixations
(TUD \cite{alers2012examining}, USC CRCNS \cite{carmi2006role}, CITIUS \cite{leboran2017dynamic}), LEDOV \cite{jiang2018deepvs}.
IRCCyN/IVC (Video 1) \cite{boulos2009region} does not mention any eye movement types at all,
while IRCCyN/IVC (Video 2) \cite{engelke2010linking} only names SP in passing.

There are two notable exceptions from this logic. First, DIEM \cite{mital2011clustering}, 
which comprises video clips from a rich spectrum of sources,
including amateur footage, TV programs and
movie trailers,
so one would expect a hugely varying fixation--pursuit balance. 
The respective paper touches on the properties of SP that separate it from fixations,
but in the end only distinguishes between blinks, saccades, and non-saccadic eye movements,
referring to the latter as generic \textit{foveations}, which combine  
fixations and SP.

GazeCom \cite{dorr2010variability}, on the other hand, explicitly acknowledges 
the difficulty of distinguishing between fixations
and smooth pursuit in dynamic scenes.
The used fixation detection algorithm
employed a dual criterion based on gaze speed and
dispersion. However, the recently published manually annotated ground truth data \cite{sp-detection-site}
shows that these coarse thresholds are insufficient to parse out SP episodes.

Part of this work's contribution is, therefore, to provide a large-scale supersaliency (SP)
and saliency (fixations) data set based on Hollywood2, as well as establishing a pipeline 
for (super)saliency evaluation.

\section{Saliency and supersaliency}

\subsection{Data sets and their analysis}
\label{sec:datasets}

\begin{figure*}[!t]
	\centering
	\subfloat[GazeCom (``street'' clip)]{\includegraphics[height=0.37\linewidth]{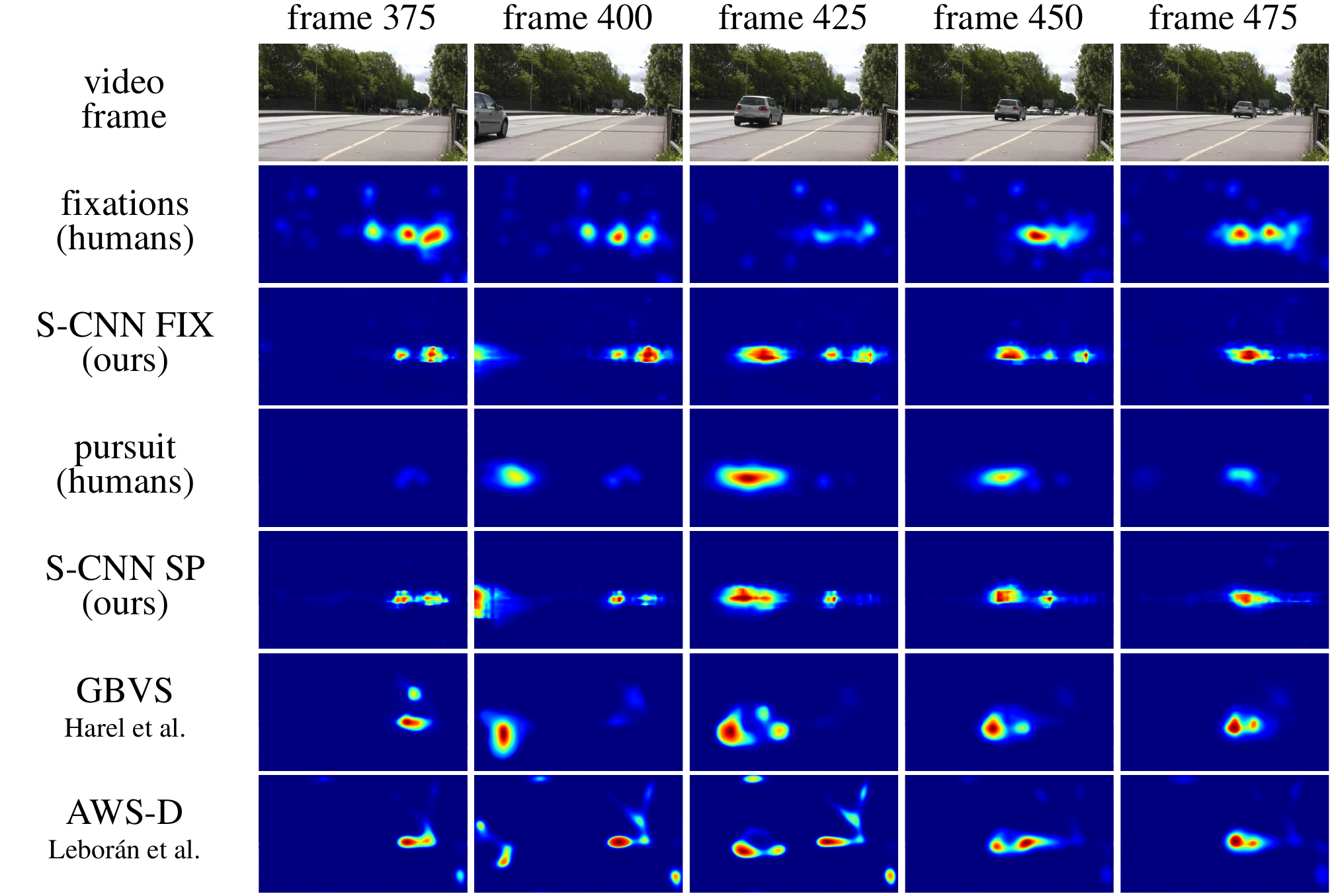}%
		\label{fig:gazecom_frames}}
	\hfill
	\subfloat[Hollywood2 (``actioncliptest00416'' clip)]{\includegraphics[height=0.37\linewidth]{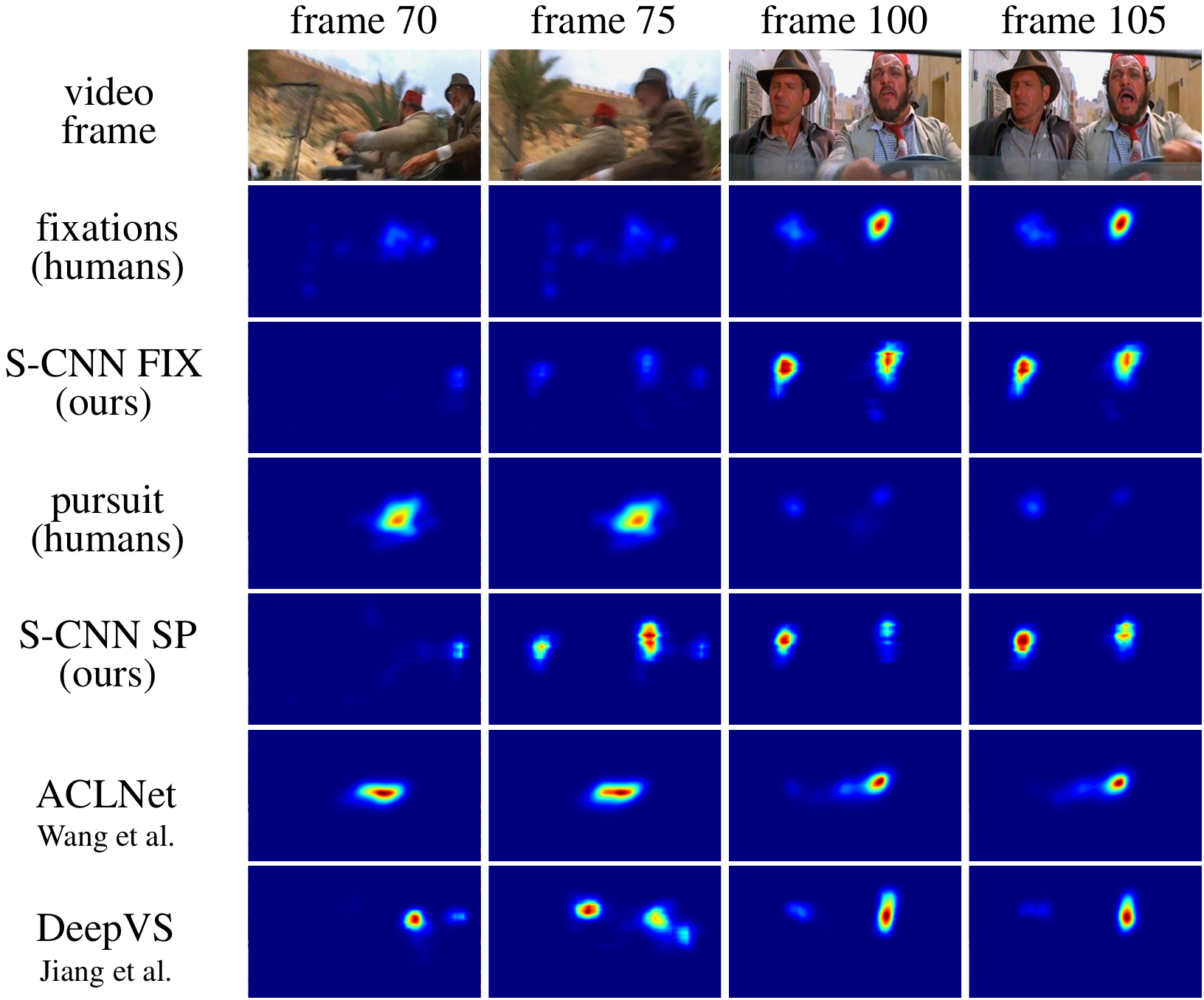}%
		\label{fig:hollywood_frames}}
	\caption{Frame examples from GazeCom (a) and Hollywood2 (b) videos (first row),
		with their respective empirical ground truth fixation-based saliency (second row) and smooth pursuit-based supersaliency 
		(fourth row) ground truth maps. Algorithmic predictions (all identically histogram-equalized, for fair visual comparison) 
		occupy the rest of the rows. The choice of saliency models 
		for visual comparison was based on best average performance on 
		the respective data set.
	}
	\label{fig:frames}
\end{figure*}

\textbf{GazeCom} \cite{dorr2010variability}, which we used because it is the only saliency data set 
that also provides full manual annotation of eye movement events \cite{agtzidis2016pursuit, sp-detection-site},
contains eye tracking data for 54 subjects,
with 18 dynamic natural scenes used as stimuli, around 20 seconds each. At over 4.5 total hours of viewing time, this is the largest manually annotated eye tracking data set that accounts for SP.
A high number of observers and the hand-labelled eye movement type information make this a suitable benchmark set. 
\figurename~{\ref{fig:gazecom_frames}} displays
an example scene, together with its empirical 
saliency maps for both fixations and smooth pursuit, and the same frames in saliency maps
predicted by different models.

\textbf{Hollywood2} \cite{mathe2012dynamic}, selected for its diversity and the sheer amount of 
eye tracking recordings,
contains about 5.5 hours of video (1707 clips,
split into training and test sets), viewed by 16 subjects. 
The movies have all types of camera movement, including translation and 
zoom, as well as scene cuts.
Here, for testing all the models, we randomly selected 50 clips from the test 
subset (same as in \figurename~{\ref{fig:sp_vs_fix:shares}}).
Example frames and respective (super)saliency maps can be seen in \figurename~{\ref{fig:hollywood_frames}}.
Since manual labelling is impractical due to the data set size, we used a publicly available toolbox \cite{sp-detection-site} implementing a state-of-the-art SP and fixation detection algorithm \cite{agtzidis2016smooth, agtzidis2016pursuit}.

\textbf{CITIUS} \cite{leboran2017dynamic} was recently used for a large-scale evaluation of the state of the art in connection with a novel model (AWS-D). 
It contains both real-life and 
synthetic video sequences, split into subcategories of static and moving camera. For our evaluation, we
used the real-life part, \mbox{\textbf{CITIUS-R}} (22 clips totalling ca.\ 7 minutes, 45 observers). 
Only fixation onset and duration data is provided,
so SP analysis was impossible.

By definition, fixations are almost stationary, so that a single point
(usually, mean gaze position placed at temporal onset) sufficiently
describes an entire fixation.
In line with the~literature, we evaluated the prediction of such fixation onsets
in the ``onset'' condition
(detected by a standard algorithm  \cite{dorr2010variability} for GazeCom and Hollywood2, provided with the data set for CITIUS-R). 
Notably, the reference models are likely optimized for this problem setting.

To describe the trajectory of an SP episode, however, all its gaze samples need to be taken into account. 
Accordingly, both the GazeCom ground truth and the toolbox \cite{sp-detection-site} we used for Hollywood2 provide sample-level annotations.
These annotations were used for prediction of individual gaze samples in the ``SP'' condition. For a direct comparison,
the ``FIX'' condition utilised individual fixation samples as well (similar to \cite{mital2011clustering}).

\subsection{Slicing CNN saliency model}

We adopted the slicing convolutional neural network (\mbox{S-CNN}) architecture \cite{shao2016slicing}, which takes an alternative approach to extracting
motion information from a video sequence. 
To achieve saliency prediction,
we extended patch-based image analysis (e.g.\ \cite{judd2009learning} for image saliency, and \cite{chaabouni2016deep}
for individual video frames) to subvolume-based video processing.
This way, we are still able to capture motion patterns, while maintaining a relatively straightforward binary classification-based architecture -- (super)salient vs.\ non-salient subvolumes. 
We do not use more complex end-to-end approaches in order to keep the proof-of-concept
implementation of fixation- and pursuit-based training as straightforward as possible,
without intermediate steps of having to convert locations of corresponding samples
into continuous saliency maps. These steps would introduce additional data 
parametrisation and, potentially, biases into the pipeline.

\begin{figure}[t!] 
	\begin{center}
		\includegraphics[width=\linewidth]{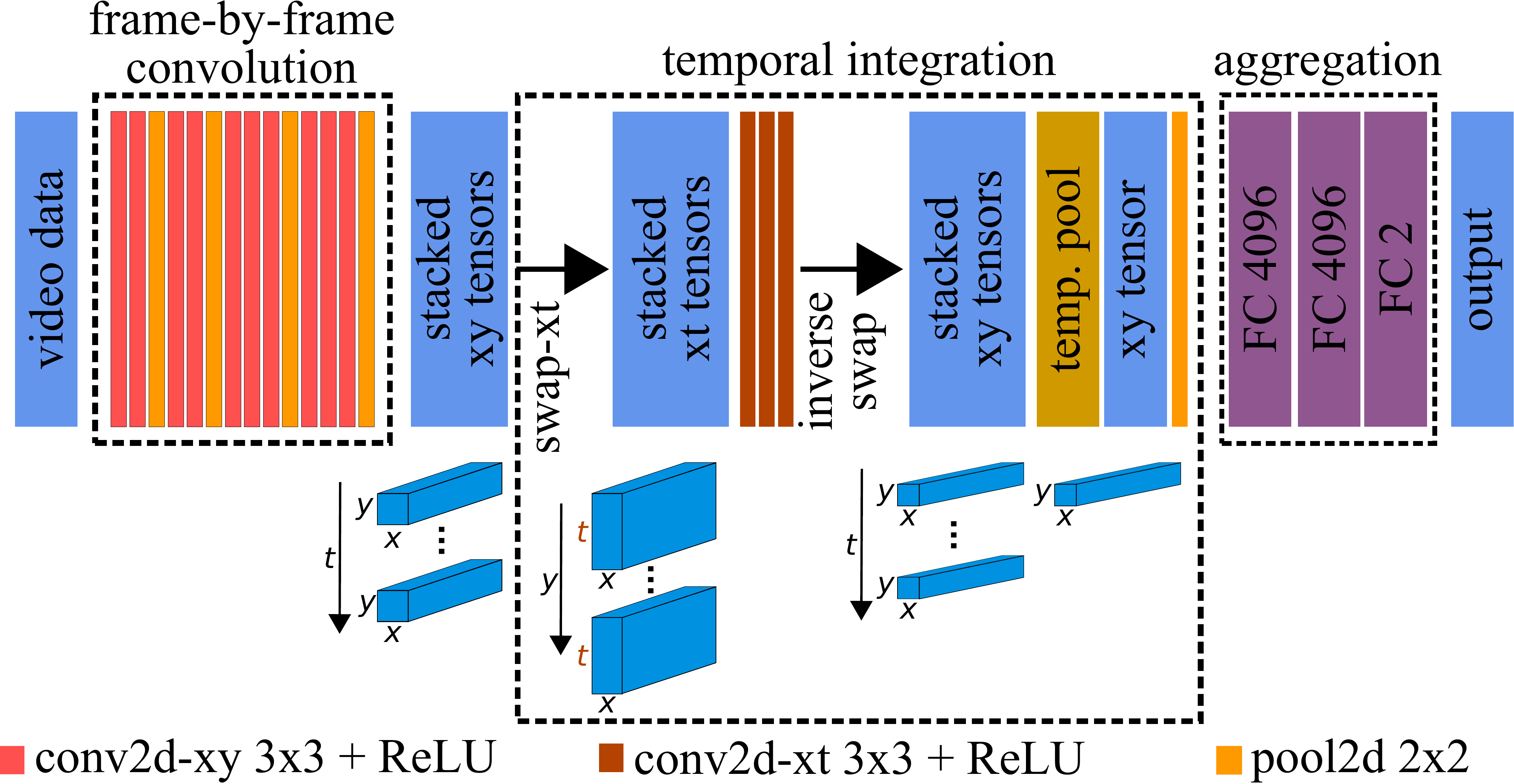}
	\end{center} 
	\caption{The $xt$ branch of the S-CNN architecture for binary salient vs.\ non-salient video subvolume classification. 
		Temporal integration is performed after the \textit{swap-xt} operation via the convolutions operating in the $xt$ plane and temporal pooling.
	} 
	\label{fig:network} 
\end{figure}

Instead of handcrafted motion descriptors \cite{chaabouni2016deep}, 3D convolutions \cite{ji2013convolutional}, or recurrent structures \cite{bazzani2017recurrent}, 
S-CNN achieves temporal integration by rotating the feature tensors after initial individual frame-based feature extraction.
This way, time (frame index) is one of the axes of the subsequent convolutions.
The architecture itself is based on VGG-16 \cite{simonyan2014deep}, with the addition of dimension swapping operations and temporal pooling.
The whole network would consist of three branches, in each of which the performed rotation is different,
and the ensuing convolutions are performed in the planes $xy$ (equivalent to no rotation), 
$xt$, or $yt$
(branches are named respectively). 
Due to the size of the complete model,
only one branch could be trained at a time. We decided to use the $xt$-branch for our experiments (see \figurename~{\ref{fig:network}}), 
since it yielded the best individual results in \cite{shao2016slicing},
and the horizontal axis seems to be more important for human vision \cite{vig2012contribution} and SP in particular \cite{rottach1996comparison}.

As input, we used RGB video subvolumes $128\texttt{\,px}\times128\texttt{\,px}\times15\texttt{\,frames}$ around the pixel to be classified. 
Similar subvolumes were used in \cite{chen2015multi} for unsupervised feature learning.  Unlike \cite{chaabouni2016deep}, we did not extract motion information explicitly, but relied on the network architecture entirely without any further input processing.

To go from binary classification to generating a continuous (super)saliency map,
we took the probability for the positive class at the soft-max layer of the network
(for each respective surrounding subvolume of each video pixel).
To reduce computation time, we only did this for every $10^{th}$ pixel along both spatial axes.
We then upscaled the resulting low-resolution map to the desired dimensions.
For GazeCom and Hollywood2, we generated saliency maps in the size $640\times360$,
whereas for CITIUS-R, the original resolution of $320\times240$ was used.

\subsection{Training details}

Out of 823 training videos in Hollywood2, 
90\% (741 clips) were used for training and 10\% for validation.
Before extracting the subvolumes centred around positive or negative
locations of our videos, these were rescaled to $640\times360$ pixels size
and mirror-padded to reduce boundary effects.
In total, the 823 clips contain
\num{4520813} unique SP and \num{10448307} unique fixated locations.
To assess the influence of eye movement type in the training data, we fitted the same model twice
for two different purposes.
First, we trained the \textit{S-CNN SP} model for predicting 
\textit{supersaliency}, so the positive locations were those
where SP had occurred. Analogously, for the \textit{S-CNN FIX} model predicting purely fixation-based (excluding SP) \textit{saliency},
the video subvolumes where observers had fixated were labelled as positive.

For both \textit{S-CNN SP} and \textit{S-CNN FIX}, the training set consisted of \num{100000} subvolumes, 
half of which were positives (as described above, randomly sampled from
the respective eye movement locations in the training videos),
half negatives (randomly selected in a uniform fashion to match the number of positive
samples per video, non-overlapping with the positive set).
For validation, \num{10000} subvolumes were used, same procedure as for the training set.

Convolutional layers were initialized 
with pre-trained VGG-16 weights, fully-connected layers were initialized randomly. 
We used a batch size of 5, and trained both models for \num{50000} iterations with 
stochastic gradient descent (with momentum of 0.9, 
learning rate starting at $10^{-4}$ and decreasing 10-fold after every \num{20000} iterations),
at which point both loss and accuracy levelled out.

\subsection{Adaptive centre bias}

Since our model is inherently spatial bias-free, as it deals purely with individual subvolumes
of the input video, we applied an adaptive solution to each frame:
The gravity centre bias approach of Wu et al.\ \cite{wu2016video},
which emphasises not the centre of the frame, but the centre of mass in the saliency distribution.
At this location, a single unit pixel is placed on the bias map, which is blurred
with a Gaussian filter ($\sigma$ equivalent to three degrees of the visual field was chosen)
and normalized to contain values ranging
from 0 to the highest saliency value of the currently processed frame.
Each frame of the video saliency map was then linearly mixed with its respective bias map (with a weight of 0.4 
for the bias, and 0.6 for the original frame, as used in~\cite{wu2016video}).

\section{Evaluation}

\subsection{Reference models}

We compared our approach to a score of publicly available dynamic saliency models. 
For compression domain models, we followed the pipeline and provided source of Khatoonabadi et al.\ \cite{khatoonabadi2015bits},
generating the saliency maps for all videos at 288 pixels in height, and proportionally scaled width
for PMES \cite{ma2001new}, MAM \cite{ma2002model}, PIM-ZEN \cite{agarwal2003fast},
PIM-MCS \cite{sinha2004region}, MCSDM \cite{liu2009motion}, MSM-SM \cite{muthuswamy2013salient},
PNSP-CS \cite{fang2014video}, and a range of OBDL-models \cite{khatoonabadi2015bits}, 
as well as pixel-domain GBVS \cite{harel2007graph, gbvs-source} and STSD \cite{haejong2009static}. 
In addition to the static AWS \cite{garciadiaz2012saliency} that was used in \cite{khatoonabadi2015bits}, we evaluated AWS-D \cite{leboran2017dynamic}, its recent extension to dynamic stimuli
(for GazeCom, after downscaling to $640\times360\,\texttt{px}$ due to memory constraints, other data sets -- at their original resolution). 
We also computed the three invariants (H, S, and K) of the structure tensor \cite{vig2012intrinsic} at fixed temporal (second) and spatial (third) scales. 
For Hollywood2, the approach of Mathe and Sminchisescu \cite{mathe2015action}, combining static (low-, mid-, and high-level) and motion features, was evaluated as well.

Deep models for saliency prediction on videos are much scarcer than such models for static images.
As of yet, the problem of finding reference models in this domain is further confounded by
the absence of publicly available code or data of some approaches, e.g.\ \cite{bazzani2017recurrent}, and
the popularity
of salient object detection approaches and data sets, e.g.\ \cite{tang2018multi, ding2018video, wang2018video}.
Included in our set of reference models are two recent approaches: 
DeepVS (OMCNN-2CLSTM) \cite{jiang2018deepvs} -- code available via \cite{deepvs-site} --
and ACLNet \cite{wang2018revisiting} -- code available via \cite{revisiting-site}.
We ran both with default parameters on all three data sets.

\subsection{Baselines}
\label{sec:baselines}

The set of baselines was inspired by the works of Judd et al.\ \cite{judd2012benchmark, mit-saliency-benchmark}:  \textit{Chance}, \textit{Permutation}, \textit{Centre},
\textit{One Human}, and \textit{Infinite Humans} (as a limit).
The latter two cannot be computed unless 
gaze data \textit{for each individual observer} are available
(i.e.\ not possible for CITIUS).
All the random baselines were repeated five times per video of each data set.
The \textit{ground truth saliency maps} were obtained via 
superimposing spatio-temporal Gaussians at every attended location of all the considered observers. 
The two spatial sigmas were set to one degree of visual angle (commonly used in the literature as the approximate fovea size, e.g.\ \cite{judd2012benchmark, salMetrics_Bylinskii}; \cite{mathe2015action} uses $1.5^\circ$). The temporal sigma was set to a frame count equivalent of $1/3$ of a second (so that the effect would be mostly contained within one second's distance).

\subsection{Metrics}
\label{sec:metrics}

\begin{table*}[!t]
	 \caption{Means and standard deviations of the absolute error of ``perfect AUC'' estimation with \emph{regular} and \emph{weighted} averaging,
	 	as well as one-sided two-sample Kolmogorov-Smirnov test p-values (with the null hypothesis that regular averaging, as a way to estimate the perfect AUC score,
	 	produces absolute errors that are smaller than or equal to those of weighted averaging). Only on CITIUS-R 
	 	we cannot say that weighted averaging has a statistically significant ($p\ll0.01$) advantage over regular averaging.}
	 \label{tab:auc-averaging}
    \begin{center}
    \begin{tabular}{|l|c|ccc|ccc|c|}
    \hline
        Statistic & Absolute error & \multicolumn{3}{c|}{GazeCom} & \multicolumn{3}{c|}{Hollywood2 (50 clips)} & \multicolumn{1}{c|}{CITIUS-R} \\
        \cline{3-9}
              & properties & SP & FIX & onsets
              & SP & FIX & onsets
              & onsets\\
        \hline 
AUC-Borji & mean (\emph{regular} averaging) & 0.038 & 0.011 & 0.012 & 0.022 & 0.017 & 0.018 & 0.0125 \\ \cline{2-9}
 & mean (\emph{weighted} averaging) & 0.011 & 0.01 & 0.01 & 0.008 & 0.012 & 0.009 & 0.0135 \\ \cline{2-9}
 & SD (\emph{regular} averaging) & 0.03 & 0.007 & 0.008 & 0.009 & 0.009 & 0.008 & 0.0079 \\ \cline{2-9}
 & SD (\emph{weighted} averaging) & 0.01 & 0.007 & 0.007 & 0.004 & 0.004 & 0.004 & 0.0075 \\ \cline{2-9}
 & \cellcolor{black!15} p-value & \cellcolor{black!15} 9e-205 & \cellcolor{black!15} 4e-16 & \cellcolor{black!15} 8e-16 & \cellcolor{black!15} 0e+00 & \cellcolor{black!15} 0e+00 & \cellcolor{black!15} 0e+00 & \cellcolor{black!15} 0.92 \\ \cline{2-9}
\cline{2-9}
\hline
sAUC & mean (\emph{regular} averaging) & 0.039 & 0.013 & 0.014 & 0.038 & 0.029 & 0.031 & 0.0173 \\ \cline{2-9}
 & mean (\emph{weighted} averaging) & 0.015 & 0.011 & 0.011 & 0.011 & 0.015 & 0.012 & 0.0169 \\ \cline{2-9}
 & SD (\emph{regular} averaging) & 0.031 & 0.008 & 0.009 & 0.016 & 0.014 & 0.013 & 0.0092 \\ \cline{2-9}
 & SD (\emph{weighted} averaging) & 0.014 & 0.008 & 0.008 & 0.006 & 0.005 & 0.005 & 0.0089 \\ \cline{2-9}
 & \cellcolor{black!15} p-value & \cellcolor{black!15} 1e-137 & \cellcolor{black!15} 4e-20 & \cellcolor{black!15} 2e-25 & \cellcolor{black!15} 0e+00 & \cellcolor{black!15} 0e+00 & \cellcolor{black!15} 0e+00 & \cellcolor{black!15} 0.02 \\ \cline{2-9}
\cline{2-9}
\hline

    \end{tabular}
    \end{center}
\end{table*}

For a thorough evaluation, we took a broad spectrum of metrics (all computed the same way for fixation samples and onsets -- saliency -- and smooth pursuit samples -- supersaliency -- for the data sets described in Section~\ref{sec:datasets}), mostly based on \cite{salMetrics_Bylinskii}: 
AUC-Judd, AUC-Borji, shuffled AUC (sAUC), 
normalized scanpath saliency (NSS),
histogram similarity (SIM), correlation coefficient (CC), and Kullback-Leibler divergence (KLD),
as well as Information Gain (IG) \cite{kummerer2015information}. We additionally computed balanced accuracy (same positive and negative location sets as for AUC-Borji; accuracy at the equal error rate point), which is the only metric that is linked to our training process almost directly.

In our implementation of sAUC and IG, in order to obtain salient locations of other clips,
we first rescaled their temporal axes
to fit the duration of the evaluated clip, and then sampled not just spatial (like e.g.\ 
\cite{leboran2017dynamic}), 
but also temporal coordinates. 
This preserves the temporal structure of the stimulus-independent bias: 
E.g.\ the first fixations after stimulus display tend to have heavier centre bias than subsequent ones in both static images \cite{tatler2007central} and videos \cite{tseng2009quantifying}.

For GazeCom and Hollywood2, we fixed all saliency maps to $640\times360\,\texttt{px}$ resolution during evaluation, either for memory constraints, or for symmetric evaluation in case of differently shaped videos. For CITIUS, the native resolution of $320\times240\,\texttt{px}$ was maintained.

\subsubsection{Metric averaging}

Due to its selectivity (i.e.\ observers can decide not to pursue anything), SP is sparse and highly unbalanced between videos (see \figurename~\ref{fig:sp_vs_fix:shares}). Using simple mean across all videos of the data set for many metrics will introduce artefacts.
For AUC-based metrics, for example,
there exists a ``perfect'' aggregated score, which
could be computed by combining the data over all the videos \textit{before} computing the metric,
i.e.\ merging all positives and all negatives beforehand.
This is, however, not always possible, as many models use per-video normalization as the final step,
either to allow for easier visualization, or to use the full spectrum of the 8-bit integer range, 
if the result is stored as a video. 
We randomly sampled non-trivial subsets of video clips 
(100 times for all the possible subset sizes)
of all three utilized test sets,
and computed per-clip AUC-Borji and sAUC scores for our \textit{S-CNN SP} model.
We combined these via either regular or weighted 
(according to the number of SP- or fixation-salient locations samples, depending on the problem setting)
averaging. This combination is then compared to the perfect score, as described above.
We found that averaging per-video
AUC scores is a significantly 
poorer approximation of the ideal score than 
their \textit{weighted mean} 
($p \ll 0.01$, for (super)saliency prediction on GazeCom and Hollywood2, see Table~\ref{tab:auc-averaging}).

We will, therefore, present the weighted averaging results for supersaliency prediction. 
Since fixations suffer from this issue to a lesser extent, this adjustment is not essential there. 
However, in the data sets with great variation of fixation samples' share
(e.g.\ Hollywood2: 30\% to 78\% in our 50-clip subset),
we would generally recommend using weighting for fixation prediction as well.
Conventional mean results for fixations are, nevertheless, presented for comparability with the literature 
(weighted results reveal a quantitatively similar picture).

\subsubsection{Cross-AUC}

Another point we raise in our evaluation is directly distinguishing SP-salient 
from fixation-salient pixels based on the saliency maps. To this end,
we introduced \textit{cross-AUC (xAUC)}: The AUC is computed for the positive 
samples' set of all pursuit-salient locations, with an equal number of 
randomly chosen fixation-salient location of the same stimulus used as negatives. 
The baselines' performance on this metric will be indicative of how well 
the targets for these two eye movements can be separated (in comparison 
to the separation of salient and non-salient locations). If a model scores 
above 50\% on this metric, it on average favours pursuit-salient locations 
over fixation-salient ones. For the purpose of distinguishing the two eye 
movement types, the scores of 70\% and 30\% are, however, equivalent.

\section{Results and discussion}

We tested the outputs of 26 published dynamic saliency models, including two deep learning-based solutions, as well as our own S-CNN models -- SP and fixation predictors both
with and without the additional post-processing step
of gravity centre bias.
For brevity and because there is no principled way of averaging different metrics numerically,
we present the results as average ranks (over the 9 metrics we used -- see Section~\ref{sec:metrics}) in Table~\ref{tab:results}.
Complete tables of all metric scores for all 7 data types (corresponding to the columns of Table~\ref{tab:results}) and 35 baselines and models can be found in the supplementary material.

\begin{table*}
\caption{Evaluation results, presented as the mean of rank values for all the metrics we compute (except for xAUC).
	``Onsets'' refers to evaluation against fixation onsets (``traditional'' saliency).
	Where marked with $^*$, ranking was computed for the weighted average of the scores.
	The rows with gray background correspond to baselines.
	Top-3 non-baseline results in each category are \textbf{boldified}.}
\label{tab:results}
\begin{center}
\begin{tabular}{|l|ccc|ccc|c|c|}
\hline
    Model & \multicolumn{3}{c|}{GazeCom} & \multicolumn{3}{c|}{Hollywood2 (50 clips)} & \multicolumn{1}{c|}{CITIUS-R} & \multicolumn{1}{c|}{{average rank}}\\
    \cline{2-8}
          & {SP$^*$} & {FIX} & {onsets}
          & {SP$^*$} & {FIX} & {onsets}
          & {onsets}
          & \\
    \hline 
\rowcolor{black!15}
    Infinite Humans & 1.0 & 1.0 & 1.0 & 1.0 & 1.0 & 1.0 & $-$ & 1.0 \\ \hline
\textit{\shortstack{S-CNN SP + Gravity CB}} & \textbf{4.9} & \textbf{2.9} & \textbf{2.9} & \textbf{4.0} & 5.1 & 5.0 & \textbf{3.3} & \textbf{4.0} \\ \hline
\textit{\shortstack{S-CNN FIX + Gravity CB}} & 12.2 & \textbf{2.8} & \textbf{2.8} & 5.3 & \textbf{4.6} & \textbf{4.1} & \textbf{3.9} & \textbf{5.1} \\ \hline
\textit{\shortstack{S-CNN SP}} & \textbf{3.0} & \textbf{4.4} & \textbf{4.1} & 6.2 & 7.6 & 7.4 & \textbf{4.8} & \textbf{5.4} \\ \hline
\textit{\shortstack{S-CNN FIX}} & \textbf{9.1} & 4.6 & 4.8 & 7.7 & 6.7 & 6.8 & 5.6 & 6.4 \\ \hline
ACLNet & 24.3 & 11.0 & 10.7 & \textbf{4.3} & \textbf{2.9} & \textbf{3.4} & \textbf{3.3} & 8.6 \\ \hline
DeepVS (OMCNN-2CLSTM) & 25.4 & 9.8 & 11.0 & \textbf{5.0} & \textbf{4.7} & \textbf{4.7} & 8.2 & 9.8 \\ \hline
GBVS & 11.1 & 11.2 & 10.1 & 11.6 & 11.3 & 11.1 & 7.6 & 10.6 \\ \hline
OBDL-MRF-O & 13.8 & 12.2 & 11.9 & 13.7 & 13.3 & 11.8 & 9.9 & 12.4 \\ \hline
OBDL-MRF-OC & 15.1 & 13.9 & 13.7 & 14.8 & 14.2 & 12.9 & 11.2 & 13.7 \\ \hline
AWS-D & 14.9 & 7.9 & 7.4 & 24.0 & 18.0 & 18.2 & 7.2 & 14.0 \\ \hline
OBDL-MRF-TO & 18.6 & 13.9 & 14.8 & 12.6 & 14.3 & 15.2 & 13.3 & 14.7 \\ \hline
OBDL-MRF & 18.9 & 16.0 & 16.3 & 13.8 & 11.3 & 12.8 & 13.7 & 14.7 \\ \hline
\rowcolor{black!15}
    Centre & 29.4 & 16.8 & 16.4 & 9.6 & 10.3 & 10.1 & 10.6 & 14.7 \\ \hline
OBDL-MRF-T & 23.1 & 13.2 & 14.9 & 12.6 & 12.2 & 15.1 & 15.3 & 15.2 \\ \hline
\rowcolor{black!15}
    One Human & 18.7 & 19.2 & 22.6 & 11.1 & 9.8 & 10.6 & $-$ & 15.3 \\ \hline
OBDL-T & 13.7 & 15.1 & 12.4 & 17.9 & 18.7 & 18.6 & 11.1 & 15.3 \\ \hline
OBDL-MRF-C & 16.3 & 16.1 & 16.0 & 15.9 & 15.9 & 14.2 & 13.1 & 15.4 \\ \hline
OBDL-MRF-TC & 20.6 & 12.9 & 14.2 & 12.8 & 16.8 & 17.1 & 15.6 & 15.7 \\ \hline
OBDL-S & 14.7 & 19.8 & 18.4 & 19.1 & 20.2 & 19.9 & 17.3 & 18.5 \\ \hline
Mathe & $-$ & $-$ & $-$ & 20.7 & 21.7 & 21.9 & $-$ & 21.4 \\ \hline
Invariant-K & 11.3 & 20.8 & 19.6 & 30.2 & 25.0 & 25.0 & 20.8 & 21.8 \\ \hline
STSD & 18.0 & 21.6 & 21.6 & 27.4 & 25.4 & 25.1 & 18.0 & 22.4 \\ \hline
OBDL & 22.4 & 23.2 & 22.4 & 22.9 & 22.9 & 22.8 & 20.8 & 22.5 \\ \hline
PMES & 11.8 & 27.8 & 27.0 & 22.0 & 27.0 & 27.4 & 27.0 & 24.3 \\ \hline
PIM-ZEN & 13.6 & 26.4 & 26.3 & 24.0 & 26.6 & 27.1 & 26.8 & 24.4 \\ \hline
PIM-MCS & 14.3 & 25.9 & 26.1 & 25.8 & 26.7 & 27.2 & 26.2 & 24.6 \\ \hline
Invariant-S & 28.6 & 21.9 & 22.2 & 32.0 & 27.7 & 27.2 & 22.8 & 26.0 \\ \hline
MSM-SM & 16.8 & 33.1 & 32.3 & 21.7 & 28.4 & 27.2 & 25.4 & 26.4 \\ \hline
PNSP-CS & 13.9 & 28.7 & 28.7 & 28.1 & 30.1 & 30.1 & 27.2 & 26.7 \\ \hline
\rowcolor{black!15}
    Permutation & 33.4 & 29.3 & 29.4 & 27.7 & 23.4 & 22.3 & 24.8 & 27.2 \\ \hline
MCSDM & 13.4 & 28.4 & 28.3 & 30.7 & 31.0 & 31.6 & 28.1 & 27.4 \\ \hline
Invariant-H & 29.8 & 23.7 & 24.4 & 33.3 & 30.9 & 30.9 & 25.8 & 28.4 \\ \hline
\rowcolor{black!15}
    Chance & 31.0 & 28.0 & 28.0 & 32.4 & 31.8 & 31.9 & 28.2 & 30.2 \\ \hline
MAM & 27.9 & 31.6 & 32.1 & 28.3 & 32.6 & 32.2 & 31.1 & 30.8 \\ \hline

\end{tabular}
\end{center}
\end{table*}

Traditional saliency prediction commonly evaluates only one sample per fixation, as we did in the ``onset'' condition.
For supersaliency, however, all gaze samples need to be predicted individually, and for consistency
we did the same for fixations in the ``FIX'' condition.
In principle, this should give greater weight to longer fixations with more samples, 
but our results show that differences between the ``FIX'' and ``onset'' conditions are small in practice (cf.\ Table~\ref{tab:results}).

On average, our pursuit prediction model, combined with adaptive centre bias (\textit{S-CNN SP + Gravity CB}), performs best, 
almost always making it to the first or the second position (and always in the top-4). Remarkably, this holds true both 
for the prediction of smooth pursuit and the prediction of fixations, 
despite training exclusively on SP-salient locations as positive examples. 
The success of our pursuit prediction approach in predicting fixations can be potentially attributed 
to humans pursuing and fixating similar targets, but the relative selectivity of SP
allows the model to focus on the particularly interesting objects in the scene.
Even without the gravity centre bias, both our saliency \textit{S-CNN FIX} 
and supersaliency \textit{S-CNN SP} models outperform the models from the literature on the whole, with their average rank 
at least ca.\ 2 positions better than that of the next best model (ACLNet).

The fact that all our \textit{S-CNN} models consistently outperform the 
traditional ``shallow'' reference 
models for both saliency and supersaliency prediction on all data sets
demonstrates the potential of deep video saliency models. This is in line with the findings in e.g.\ \cite{bazzani2017recurrent,bak2017spatio}, where a deep architecture has shown superior fixation detection performance, compared to non-CNN models. 
On Hollywood2, due to the very centre biased nature of the gaze locations \cite{dorr2010variability}, for example, only the deep learning models 
(\emph{S-CNN}, ACLNet, and DeepVS)
rank higher than the Centre Baseline or achieve non-negative information gain scores (cf.\ 
Table~\ref{tab:results} and the tables in the supplementary material).

Only in the fixation prediction task on the 
Hollywood2 data set, the results of our best model are inferior to the two deep reference approaches
(and only to those) -- DeepVS and ACLNet.  On both other data sets (GazeCom and CITIUS-R), as well as for supersaliency
prediction on Hollywood2, this model is outperforming all reference algorithms.
The two evaluated deep literature approaches are particularly weak on the GazeCom 
data set, and especially in the task of predicting pursuit-based supersaliency. 
Qualitatively, we observed that their predicted saliency distributions tend to miss moving salient targets, 
unless these are close to the centre of the frame.

Both with and without the gravity centre bias, our supersaliency \textit{S-CNN SP} models perform better than our respective saliency \textit{S-CNN FIX} models 
(with the difference in average rank value of ca.\ one position). 
We emphasise that these models are only trained on the Hollywood2 training set. 
On the Hollywood2 test set, maybe not surprisingly, the fixation-predicting models perform better for fixation-based saliency and 
SP-predicting models perform better for pursuit-based supersaliency. 
On the two other
data sets, however, the models that were trained for SP prediction generally perform better than their fixation-trained counterparts, indicating their 
greater generalization capability.

To find informative video regions, we use humans as a yardstick, 
since they clearly excel at real-world tasks despite their limited perceptual throughput.
Smooth pursuit is more selective than fixations and thus 
likely restricted to particularly interesting objects.
The use of such sparser, higher-quality training data could explain
the superior generalizability of the supersaliency models to independent data sets.

For visual comparison, example saliency map sequences are presented in \figurename s~\ref{fig:gazecom_frames} and \ref{fig:hollywood_frames} for select
GazeCom and Hollywood2 clips, respectively. It can be seen, for example, that our \textit{S-CNN FIX} model differentiates well between fixation-rich and SP-rich frames in the Hollywood2 clip.

\subsection{Distinguishing fixation and pursuit targets}

In the task of separating SP- and fixation-salient locations (the xAUC metric),
most models yield a result above 0.5 on GazeCom, which means that they
still, by chance or by design, assign higher saliency values to SP locations (unlike e.g.\ the 
centre baseline with xAUC score of 0.44, which implies that fixations
on this data set are more centre biased than pursuit). 
Probably due to their emphasis on motion information, the top of the chart with respect to this metric
is heavily dominated by compression-domain approaches (top-7 non-baseline models for GazeCom, top-4 for Hollywood2, cf.\ tables in the supplementary material).
Even though in the limit (\textit{Infinite Humans} baseline) this metric's weighted average can be 
confidently above 0.9, the best model's (MSM-SM \cite{muthuswamy2013salient}) 
result is just below 0.74 for GazeCom, and below 0.6 for Hollywood2 . 
This particular aspect needs more investigation and, possibly, dedicated training.

\section{Conclusion}

In this paper, we introduced the concept of \textit{supersaliency} -- smooth pursuit-based attention prediction. We argue that pursuit exhibits properties that set it apart from fixations in terms of perception and behavioural consequences, and that predicting smooth pursuit should thus be studied separately from fixation prediction.
To this end, we developed a novel pipeline and tested it on the ground truth for saliency and supersaliency problems for the large-scale Hollywood2, as well as for a manually annotated GazeCom.

To better understand a model's behaviour on supersaliency data, we
introduced the cross-AUC metric that assesses an algorithm's preference for pursuit
vs.\ fixation locations, thus describing its ability to distinguish between the
two.  While the human data showed that there are clear systematic differences
between the two target types, it remains an open question how to reliably capture these differences
with video-based saliency models.

Finally, we proposed and evaluated a deep saliency model with the slicing CNN architecture, which we trained for both smooth pursuit and fixation-based attention prediction.
In both settings, our model outperformed all 26 tested dynamic reference models (on average). Importantly, training for supersaliency yielded better results even for traditional fixation-based saliency prediction  on two additional independent data sets, i.e.\ supersaliency showed better generalizability. These findings demonstrate the potential of smooth pursuit modelling and prediction. 

\section*{Acknowledgements}
Supported by the Elite Network Bavaria, funded by the
Bavarian State Ministry for Research and Education.


{\small 
\bibliographystyle{ieee} 
\bibliography{paper} }

\end{document}